\documentclass{ecai}
\usepackage{times}
\usepackage{tabularx}

\usepackage{graphicx}
\usepackage{latexsym}
\usepackage{amsmath}

\begin{document}

\title{Curriculum Learning with Diversity \\for Supervised Computer Vision Tasks}

\author{Petru Soviany\institute{University of Bucharest, Department of Computer Science,
Romania, email: petru.soviany@yahoo.com}}

\maketitle
\bibliographystyle{ecai}

\begin{abstract}
  Curriculum learning techniques are a viable solution for improving the accuracy of automatic models, by replacing the traditional random training with an easy-to-hard strategy. However, the standard curriculum methodology does not automatically provide improved results, but it is constrained by multiple elements like the data distribution or the proposed model. In this paper, we introduce a novel curriculum sampling strategy which takes into consideration the diversity of the training data together with the difficulty of the inputs. We determine the difficulty using a state-of-the-art estimator based on the human time required for solving a visual search task. We consider this kind of difficulty metric to be better suited for solving general problems, as it is not based on certain task-dependent elements, but more on the context of each image. We ensure the diversity during training, giving higher priority to elements from less visited classes. We conduct object detection and instance segmentation experiments on Pascal VOC 2007 and Cityscapes data sets, surpassing both the randomly-trained baseline and the standard curriculum approach. We prove that our strategy is very efficient for unbalanced data sets, leading to faster convergence and more accurate results, when other curriculum-based strategies fail.
\end{abstract}

\section{Introduction}
\vspace{0.3cm}
Although the accuracy of automatic models highly increased with the development of deep and very deep neural networks, an important and less studied key element for the overall performance is the training strategy. In this regard, Bengio et al.~\cite{Bengio-ICML-2009} introduced curriculum learning (CL), a set of learning strategies inspired by the way in which humans teach and learn. People learn the easiest concepts at first, followed by more and more complex elements. Similarly, CL uses the difficulty context, feeding the automatic model with easier samples at the beginning of the training, and gradually adding more difficult data as the training proceeds. 

The idea is straightforward, but an important question is how to determine whether a sample is easy or hard. CL requires the existence of a predefined metric which can compute the difficulty of the input examples. Still, the difficulty of an image is strongly related to the context: a big car in the middle of an empty street should be easier to detect than a small car, parked in the corner of an alley full of pedestrians. Instead of building hand-crafted models for retrieving contextual information, in this paper, we use the image difficulty estimator from~\cite{Ionescu-CVPR-2016} which is based on the amount of time required by human annotators to assess if a class is present or not in a certain image. We consider that people can understand the full context very accurately, and that a difficulty measure trained on this information can be useful in our setting.

The next challenge is building the curriculum schedule, or the rate at which we can augment the training set with more complex information. To address this problem, we follow a sampling strategy similar to the one introduced in~\cite{Soviany_2020_WACV}. Based on the difficulty score, we sample according to a probability function, which favors easier samples in the first iterations, but converges to give the same weight to all the examples in the later phases of the training. Still, the probability of sampling a harder example in the first iterations is not null, and the more difficult samples which are occasionally picked increase the diversity of the data and help training.

The above-mentioned methodology should work well for balanced data sets, as various curriculum sampling strategies have been successfully employed in literature~\cite{liu2018curriculum,Soviany_2020_WACV,Wang_2019_ICCV,zhang2018empirical}, but it can fail when the data is unbalanced. Ionescu et al.~\cite{Ionescu-CVPR-2016} show that some classes may be more difficult than others. A simple motivation for this may be the context in which each class appears. For example, a potted plant or a bottle are rarely the focus of attention, usually being placed somewhere in the background. Other classes of objects, such as tables, are usually occluded, with the pictures focusing on the objects on the table rather than on the piece of furniture itself. This can make a standard curriculum sampling strategy neglect examples from certain classes and slow down training. The problem becomes even more serious in a context where the data is biased towards the easier classes. To solve these issues, we add a new term to our sampling function which takes into consideration the classes of the elements already sampled, in order to emphasize on images from less-visited classes and ensure the diversity of the selected examples.

The importance of diversity can be easily explained when comparing our machine learning approach to actual real-life examples. For instance, when creating a new vaccine, researchers need to experiment on multiple variants of the virus, then test it on a diverse group of people. As a rule, in all sciences, before making any assumptions, researchers have to examine a diverse set of examples which are relevant to the actual data distribution. Similar to the vaccines, which must be efficient for as many people as possible, we want our curriculum model to work well on all object classes. We argue that this is not possible in unbalanced curriculum scenarios, and it is slower in the traditional random training setup.

Since it is a sampling procedure, our CL approach can be applied to any supervised task in machine learning. In this paper, we focus on object detection and instance segmentation, two of the main tasks in computer vision, which require the model to identify the class and the location of objects in images. To test the validity of our approach, we experiment on two data sets: Pascal VOC 2007~\cite{pascal-voc-2007} and Cityscapes~\cite{Cordts2016Cityscapes}, and compare our curriculum with diversity strategy against the standard random training method, a curriculum sampling (without diversity) procedure and an inverse-curriculum approach, which selects images from hard to easy. We employ a state-of-the-art Faster R-CNN~\cite{ren2015faster} detector with a Resnet-101~\cite{he2016deep} backbone for the object detection experiments, and a Mask R-CNN~\cite{he2017mask} model based on Resnet-50 for instance segmentation.

Our main contributions can be summarized as follows: 

\vspace*{-0.28cm}\begin{enumerate}
   \item We illustrate the necessity of adding diversity when using CL in unbalanced data sets; 
   \item We introduce a novel curriculum sampling function, which takes into consideration the class-diversity of the training samples and improves results when traditional curriculum approaches fail; 
   \item We prove our strategy by experimenting on two computer vision tasks: object detection and instance segmentation, using two data sets of high interest.
\end{enumerate}

\vspace*{-0.28cm}We organize the rest of this paper as follows: in Section~\ref{sec_RelatedWork}, we present the most relevant related works and compare them with our approach. In Section~\ref{sec_Method}, we explain in detail the methodology we follow. We present our results in Section~\ref{sec_Experiments}, and draw our conclusion and discuss possible future work in the last section.

\section{Related Work}
\label{sec_RelatedWork}

{\bfseries Curriculum learning.} Bengio et al.~\cite{Bengio-ICML-2009} introduced the idea of curriculum learning (CL) to train artificial intelligence, proving that the standard learning paradigm used in human educational systems could also be applied to automatic models. CL represents a class of easy-to-hard approaches, which have successfully been employed in a wide range of machine learning applications, from natural language processing~\cite{guo2019finetuning,kocmi2017curriculum,liu2018curriculum,platanios-etal-2019-competence,subramanian2017adversarial}, to computer vision~\cite{gonga2016,gui2017,Hacohen2019OnTP,jiang2018mentornet,li17,shi2016weakly,Weinshall2018CurriculumLB}, or audio processing~\cite{Amodei2016,ranjan2017curriculum}. 

One of the main limitations of CL is that it assumes the existence of a predefined metric which can rank the samples from easy to hard. These metrics are usually task-dependent with various solutions being proposed for each. For example, in text processing, the length of the sentence can be used to estimate the difficulty of the input (shorter sentences are easier)~\cite{platanios-etal-2019-competence,spitkovsky2009babysteps}, while the number and the size of objects in a certain sample can provide enough insights about difficulty in image processing tasks (images with few large objects are easier)~\cite{shi2016weakly,Soviany-CEFRL-2018}. In our paper, we employ the image difficulty estimator of Ionescu et al.~\cite{Ionescu-CVPR-2016} which was trained considering the time required by human annotators to identify the presence of certain classes in images.

To alleviate the challenge of finding a predefined difficulty metric, Kumar et al.~\cite{kumar2010self} introduce self-paced learning (SPL), a set of approaches in which the model ranks the samples from easy to hard during training, based on its current progress. For example, the inputs with the smaller loss at a certain time during training are easier than the samples with higher loss. Many papers apply SPL successfully~\cite{sangineto2018self,supancic2013self,tang2012shifting}, and some methods combine prior knowledge with live training information, creating self-paced with curriculum techniques~\cite{jiang2015self,zhang2019leveraging}. Even so, SPL still has some limitations, requiring a methodology on how to select the samples and how much to emphasize easier examples. Our approach is on the borderline between CL and SPL, but we consider it to be pure curriculum, although we use training information to advantage less visited classes. During training, we only count the labels of the training samples, which is a priori information, and not the learning progress. A similar system could iteratively select examples from every class, but this would force our model to process the same number of examples from each class. Instead, by using the class-diversity as a term in our difficulty-based sampling probability function, we impose the selection of easy-to-hard diverse examples, without massively altering the actual class distribution of the data set.  

The easy-to-hard idea behind CL can be implemented in multiple ways. One option is to start training on the easiest set of images, while gradually adding more difficult batches~\cite{Bengio-ICML-2009,gui2017,kocmi2017curriculum,shi2016weakly,spitkovsky2009babysteps,zhang2018empirical}. Although most of the models keep the visited examples in the training set, Kocmi et al.~\cite{kocmi2017curriculum} suggest reducing the size of each bin until combining it with the following one, in order to use each example only once during an epoch. In~\cite{liu2018curriculum,Soviany_2020_WACV} the authors propose a sampling strategy according to some probability function, which favors easier examples in the first iterations. As the authors show, the easiness score from~\cite{Soviany_2020_WACV} could also be added as a new term to the loss function to emphasize the easier examples in the beginning of the training. In this paper, we enhance their sampling strategy by adding a new diversity term to the probability function used to select training examples.

\begin{figure}
\begin{center}
\centerline{\includegraphics[width=\columnwidth]{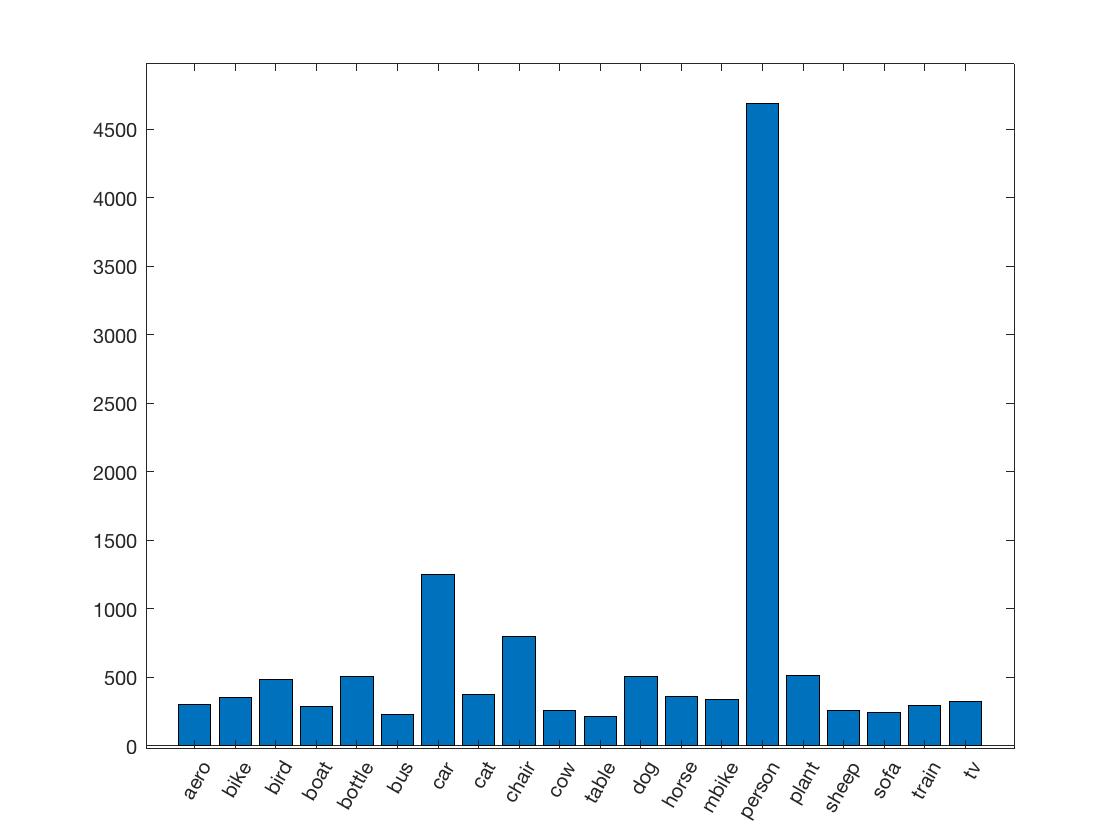}}
\caption{Number of instances from each class in the trainval split of the Pascal VOC 2007 data set.}
\label{distrib}
\end{center}
\end{figure}

Despite leading to good results in many related papers, the standard CL procedure is highly influenced by the task and the data distribution. Simple tasks may not gain much from using curriculum approaches, while employing CL in unbalanced data sets can lead to slower convergence. To address the second problem, Wang et al.~\cite{Wang_2019_ICCV} introduce a CL framework which adaptively adjusts the sampling strategy and loss weight in each batch, while other papers~\cite{jiang2014self,sachan2016easy} argue that a key element is diversity. Jiang et al.~\cite{jiang2014self} introduce a SPL with diversity technique in which they regularize the model using both difficulty information and the variety of the samples. They suggest using clustering algorithms to split the data into diverse groups. Sachan et al.~\cite{sachan2016easy} measure diversity using the angle between the hyperplanes the samples induce in the feature
space. They choose the examples that optimize
a convex combination of the curriculum learning
objective and the sum of angles between the candidate samples and the examples selected in previous steps. In our model, we define diversity based on the classes of our data. We combine our predefined difficulty metric with a score which favors images from less visited classes, in order to sample easy and diverse examples at the beginning of the training, then gradually add more complex elements. Our idea works well for supervised tasks, but it can be extended to unsupervised learning by replacing the ground-truth labels with a clustering model, as suggested in~\cite{jiang2014self}. Figure~\ref{distrib} presents the class distribution on Pascal VOC 2007 data set~\cite{pascal-voc-2007} which is heavily biased towards class \emph{person}.

\begin{figure*}[ht]
\begin{center}
\centerline{\includegraphics[width=\linewidth]{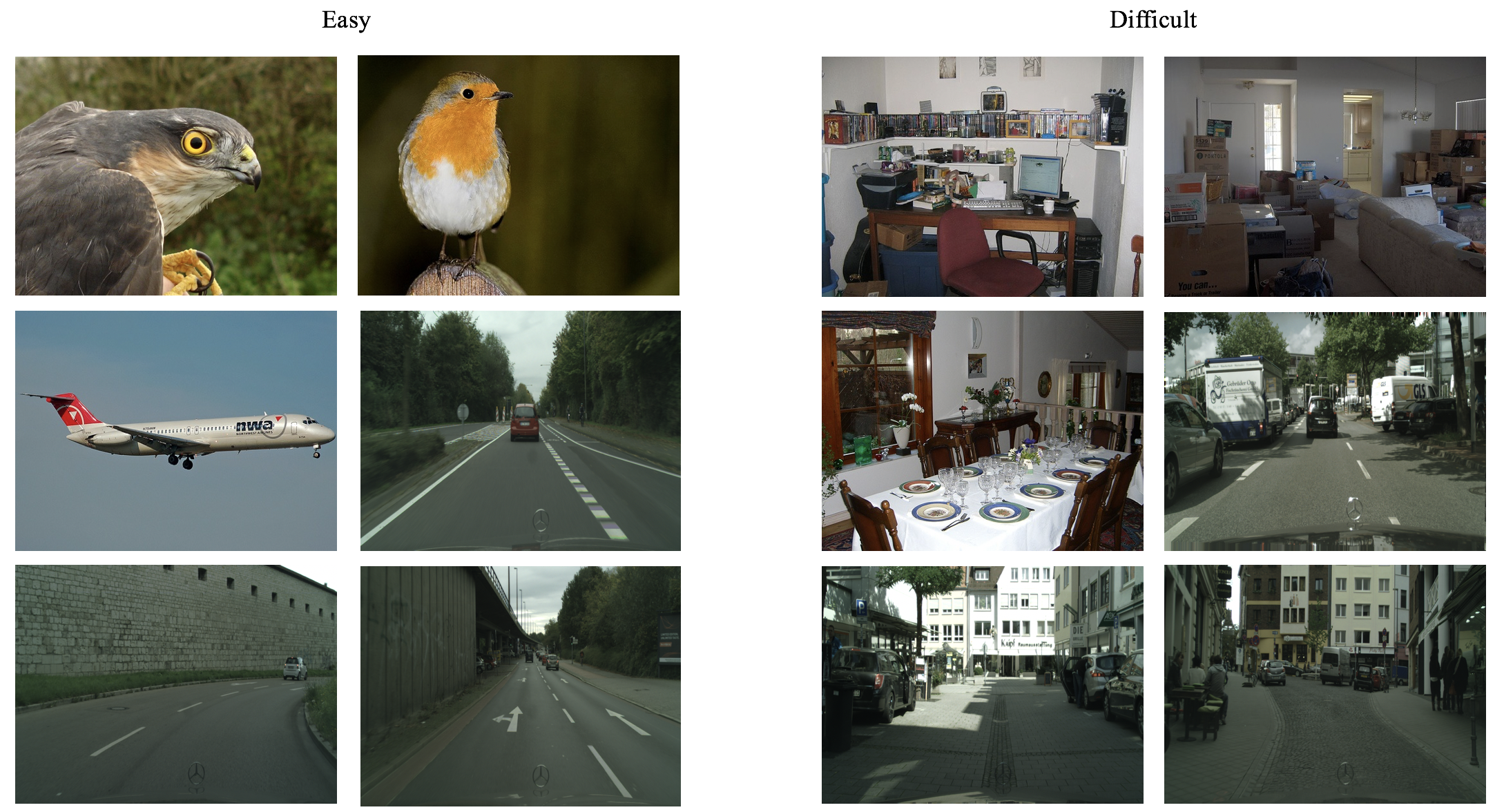}}
\caption{Easy and difficult images from Pascal VOC 2007 and Cityscapes according to our estimation.}
\label{diff}
\end{center}
\end{figure*}
{\bfseries Object detection} is the task of predicting the location and the class of objects in certain images. As noted in~\cite{Soviany-CEFRL-2018}, the state-of-the-art object detectors can be split into two main categories: two-stage and single stage models. The two-stage object detectors~\cite{he2017mask,ren2015faster} use a Region Proposal Network to generate regions of interest which are then fed to another network for object localization and classification. The single stage approaches~\cite{liu2016ssd,redmon2016} take the whole image as input and solve the problem like a regular regression task. These methods are usually faster, but less accurate than the two-stage designs. {\bfseries Instance segmentation} is similar to object detection, but more complex, requiring the generation of a mask instead of a bounding box for the objects in the test image. Our strategy can be implemented using any detection and segmentation models, but, in order to increase the relevance of our results, we experiment with high quality Faster R-CNN~\cite{ren2015faster} and Mask R-CNN~\cite{he2017mask} baselines.
\section{Methodology}
\label{sec_Method}
Training artificial intelligence using curriculum approaches, from easy to hard, can lead to improved results in a wide range of tasks~\cite{Amodei2016,gonga2016,gui2017,guo2019finetuning,Hacohen2019OnTP,jiang2018mentornet,kocmi2017curriculum,li17,liu2018curriculum,platanios-etal-2019-competence,ranjan2017curriculum,shi2016weakly,subramanian2017adversarial,Weinshall2018CurriculumLB}. Still, it is not simple to determine which samples are easy or hard, and the available metrics are usually task-dependent. Another challenge of CL is finding the right curriculum schedule, i.e. how fast to add more difficult examples to training, and how to introduce the right amount of harder samples at the right time to positively influence convergence. In this section, we present our approach for estimating difficulty and our curriculum sampling strategies.

\subsection{Difficulty estimation}
To estimate the difficulty of our training examples, we employ the method of Ionescu et al.~\cite{Ionescu-CVPR-2016} who defined image difficulty as the human time required for solving a visual search task. They collected annotations for the Pascal VOC 2012~\cite{pascal-voc-2012} data set, by asking annotators whether a class was present or not in a certain image. They collected the time people required for answering these questions, which they normalized and fed as training data for a regression model. Their results correlate fine with other difficulty metrics which take into consideration the number of objects, the size of the objects, or the occlusions. Because it is based on human annotations, this method takes into account the whole image context, not only certain features relevant for one problem (the number of objects, for example). This makes the model task independent, and, as a result, it was successfully employed in multiple vision problems~\cite{Ionescu-CVPR-2016,Soviany-CEFRL-2018,Soviany_2020_WACV}. To further prove the efficiency of the estimator for our task, we show that automatic models have a lower accuracy in difficult examples. We split the Pascal VOC 2007~\cite{pascal-voc-2007} test set in three equal batches: easy, medium and hard, and run the baseline model on each of them. The results in Table~\ref{tab10} confirm that the AP lowers as the difficulty increases.

We follow the strategy of Ionescu et al. as described in the original paper~\cite{Ionescu-CVPR-2016} to determine the difficulty scores of the images in our data sets. These scores have values $\approx{3}$, with a larger score defining a more difficult sample. We translate the values between $[-1,1]$ using Equation~\ref{minmax} to simplify the usage of the score in the next steps. Figure~\ref{diff} shows some examples of easy and difficult images.
\begin{equation}
\label{minmax}
Scale_{min-max}(x) = \frac{2 \cdot (x - min(x))}{max(x)-min(x)} - 1
\end{equation}

\begin{table}
\caption{Average Precision scores for object detection using the baseline Faster R-CNN, on easy, medium and hard splits of Pascal VOC 2007 test set, as estimated using our approach.}\label{tab10}
\centering
\begin{tabularx}{7cm}{| *2{>{\centering\arraybackslash}X}|}
\hline
DIfficulty & mAP (in \%)\\
\hline
Easy & $72.93$\\
Medium & $72.16$\\
Hard & $67.03$\\
\hline
\end{tabularx}
\end{table}

\subsection{Curriculum sampling}
Soviany et al.~\cite{Soviany_2020_WACV} introduce a curriculum sampling strategy, which favors easier examples in the first iterations and converges as the training progresses. It has the advantage of being a continuous method, removing the necessity of a curriculum schedule for enhancing the difficulty-based batches. Furthermore, the fact that it is a probabilistic sampling method does not constrain the model to only select easy examples in the first iterations, as batching does, but adds more diversity in data selection. We follow their approach in building our curriculum sampling strategy with only a small change in the position of parameter $k$ in order to better emphasize the difficulty of the examples. We use the following function to assign weights to the input images during training: 
\begin{equation}
\label{scoring}
w(x_i,t) = {(1 - diff(x_i) \cdot e^{-\gamma \cdot t})}^k , \forall x_i \in X,
\end{equation}
where $x_i$ is the training example from the data set X, $t$ is the current iteration, and $diff(x_i)$ is the difficulty score associated with the selected sample. $\gamma$ is a parameter which sets how fast the function converges to $1$, while $k$ sets how much to emphasize the easier examples. Our function varies from the one proposed in~\cite{Soviany_2020_WACV} by changing the position of the $k$ parameter. We consider that we can take advantage of the properties of the power function which increases faster for numbers greater than the unit. Since $1 - s_i \cdot e^{-\gamma \cdot t} \in [0,2]$, and the result is $> 1$ for easier examples, our function will focus more on the easier samples in the first iterations. As the training advances, the function converges to $1$, so all examples will have the same probability to be selected in the later phases of the training. We transform the weights into probabilities and we sample accordingly.

\subsection{Curriculum with diversity sampling}
As~\cite{jiang2014self,sachan2016easy} note, applying a CL strategy does not guarantee improved quality, the diversity of the selected samples having a great impact on the final results. A simple example is the case in which the data set is biased, having fewer samples of certain classes. Since some classes are more difficult than others~\cite{Ionescu-CVPR-2016}, if the data set is not well-balanced, the model will not visit the harder classes until the later stages of the training. Thus, the model will not perform well on classes it did not visit. This fact is generally valid in all kind of applications, even in real life reasoning: without seeing examples which match the whole data distribution, it is impossible to find the solution suited for all scenarios. Because of this, we enhance our sampling method, by adding a new term, which is based on the diversity of the examples.

Our diversity scoring algorithm is simple, taking into consideration the classes of the selected samples. During training, we count the number of visited objects from each class ($num_{objects}(c)$). We subtract the mean of the values to determine how often each class was visited. This is formally presented in Equation~\ref{visited}. We scale and translate the results between $[-1,1]$ using Equation~\ref{minmax} to get the score of each class, then, for every image, we compute the image-level diversity by averaging the class score for each object in its ground-truth labels (Equation~\ref{imgvisited}).

\begin{multline}\label{visited}
\shoveright{visited(c_i) = num_{objects}(c_i) - \frac{\sum_{c_j \in C} num_{objects}(c_j)}{|C|}}\\
\forall c_i \in C.
\end{multline}

\noindent\begin{multline}
\label{imgvisited}
\shoveright{imgVisited(x_i) = \frac{\sum_{obj \in objects(x_i)} visited(class(obj))}{|objects(x_i)|}}\\\forall x_i \in X.
\end{multline}

\vspace*{0.3cm}In our diversity algorithm we want to emphasize the images containing objects from less visited classes, i.e. with a small $imgVisited$ value, closer to $-1$. We compute a scoring function similar to Equation~\ref{scoring}, which also takes into consideration how often a class was visited, in order to add diversity:
\begin{multline}
\label{samplingfinal}
w(x_i,t) = [1 - \alpha \cdot (diff(x_i) \cdot e^{-\gamma \cdot t}) 
\\
- (1 - \alpha) \cdot (imgVisited(x_i) \cdot e^{-\gamma \cdot t})]^k,
\end{multline}
where $\alpha$ controls the impact of each component, the difficulty and the diversity, while the rest of the notation follows Equation~\ref{scoring}. We transform the weights into probabilities by dividing them by their sum, and we sample accordingly.

\begin{figure}[ht]

\begin{center}
\centerline{\includegraphics[width=\columnwidth]{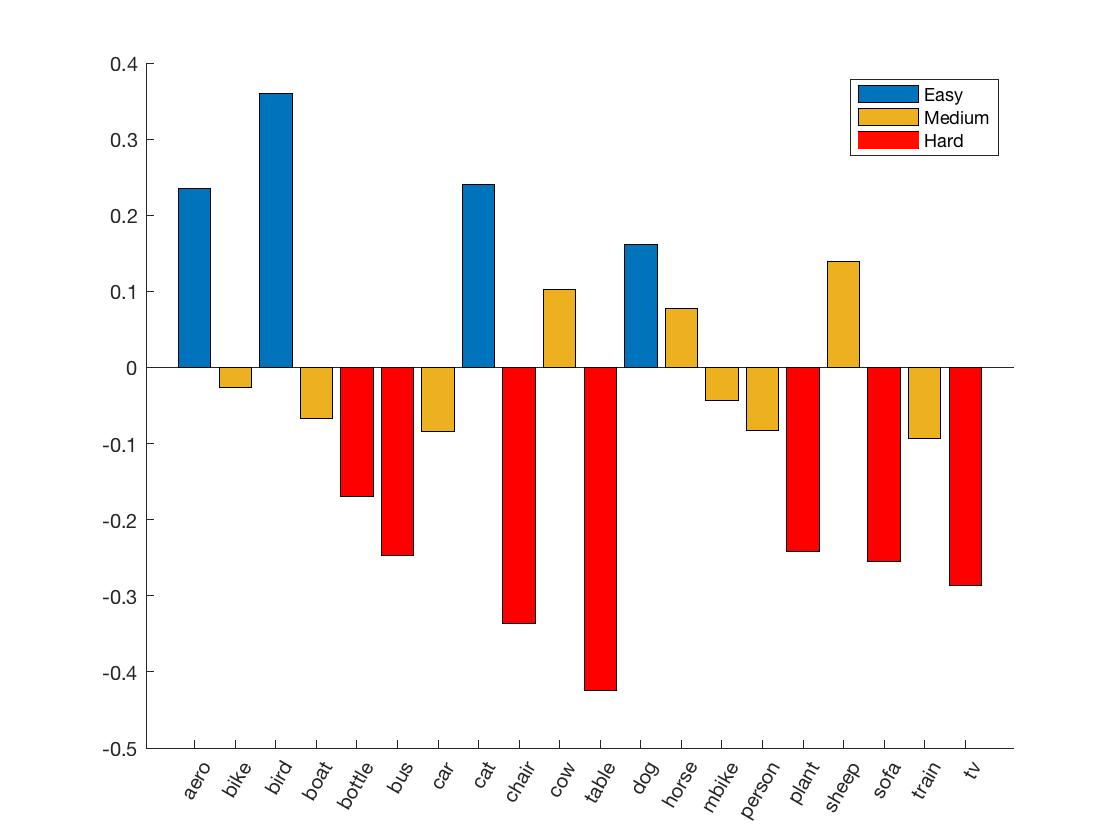}}
\caption{Difficulty of classes in Pascal VOC 2007 according to our estimation. Best viewed in color.}
\label{diff2}
\end{center}
\end{figure}

\begin{figure*}[t]
\begin{center}
\includegraphics[width=0.94\linewidth]{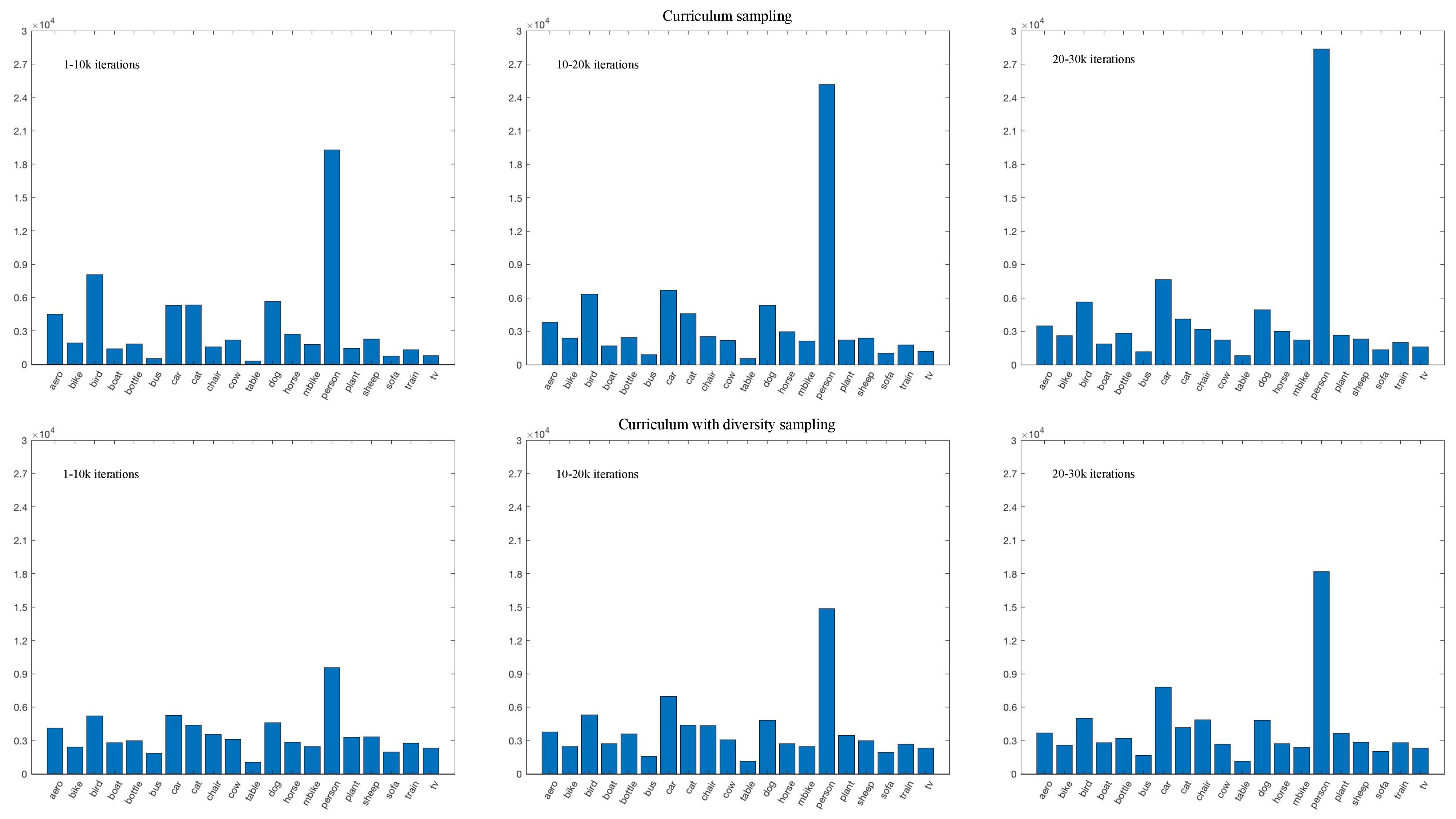}
\end{center}
 \caption{Number of objects from each class sampled during our training on Pascal VOC 2007. On the first row it is the curriculum sampling method and on the second row it is the curriculum with diversity approach. We present the first 30000 iterations for each case, with histograms generated from 10k to 10k steps.}
 \label{fig1}
\end{figure*}

\section{Experiments}\label{sec_Experiments}
\subsection{Data sets}
In order to test the validity of our method, we experiment on two data sets: Pascal VOC 2007~\cite{pascal-voc-2007} and Cityscapes~\cite{Cordts2016Cityscapes}. We conduct detection experiments on 20 classes, training on the 5011 images from the Pascal VOC 2007 trainval split. We perform evaluation on the test split which contains 4952 images. For our instance segmentation experiments, we use the Cityscapes data set which contains eight labeled object classes: person, rider, car, truck, bus, train, motorcycle, bicycle. We train on the training set of 2975 images and we evaluate on the validation split of 500 images. 
\subsection{Baselines and configuration}
We build our method on top of the Faster R-CNN~\cite{ren2015faster} and Mask R-CNN~\cite{he2017mask} implementations available at:  https://github.com/facebookresearch/maskrcnn-benchmark.
For our detection experiments, we use Faster R-CNN with Resnet-101~\cite{he2016deep} backbone, while for segmentation we employ the Resnet-50 backbone on the Mask R-CNN model. We use the configurations available on the web site, with the learning rate adjusted for a training with a batch size of 4. In our sampling procedure (Equation~\ref{samplingfinal}) we set $\alpha = 0.5$, $\gamma = 6 \cdot 10^{-5}$, and $k=5$. We do not compare with other models, because the goal of our paper is not surpassing the state of the art, but improving the quality of our baseline model. We also present the results of a hard-to-easy sampling, in order to prove the efficiency of the easy-to-hard curriculum approaches inspired by human learning.

\begin{figure}[ht]
\vskip 0.2in
\begin{center}
\centerline{\includegraphics[width=\columnwidth]{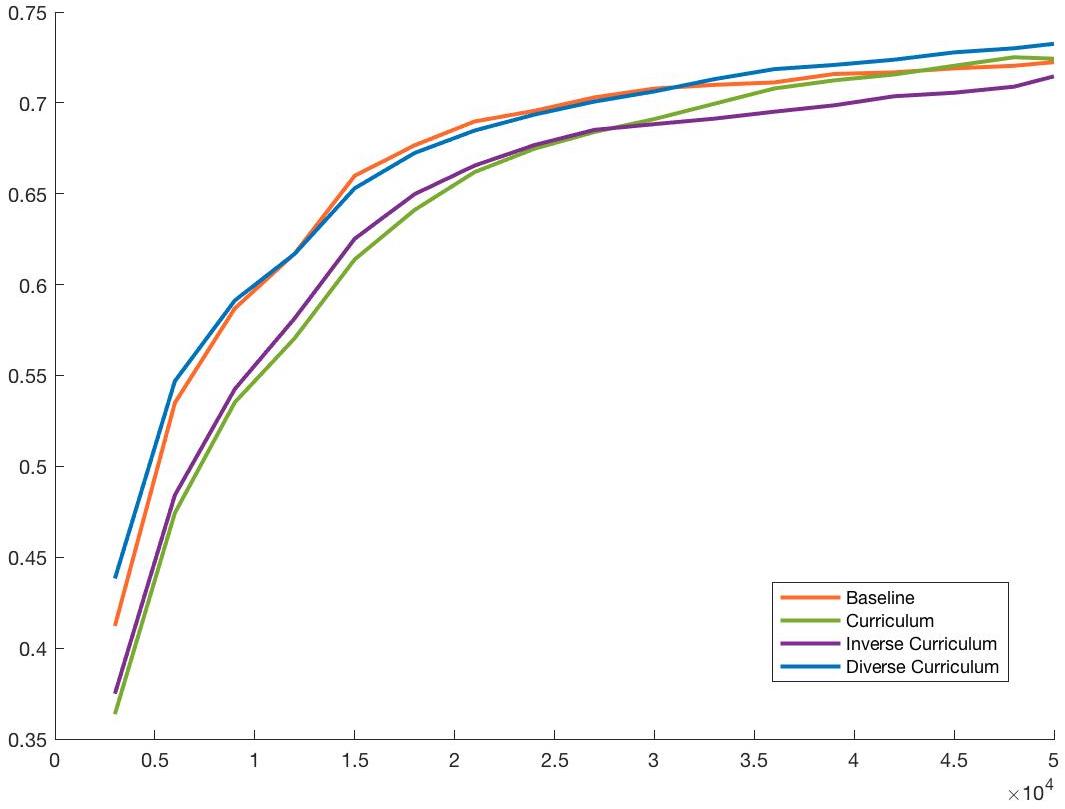}}
\caption{Evolution of mAP during training on Pascal VOC 2007 for object detection. Best viewed in color.}
\label{fig3}
\end{center}
\vskip -0.2in
\end{figure}

\subsection{Evaluation metrics}
We evaluate our results using the mean Average Precision (AP). The AP score is given by the area under the precision-recall curve for the detected objects. The Pascal VOC 2007~\cite{pascal-voc-2007} metric is the mean of precision values at a set of 11 equally spaced recall levels, from 0 to 1, at a step size of 0.1. The Cityscapes~\cite{Cordts2016Cityscapes} metric computes the average precision on the region level for each class and averages it across 10 different overlaps ranging from 0.5 to 0.95 in steps of 0.05. We also report results on Cityscapes using AP50\% and AP75\%, which correspond to overlap values of 50\% and 75\%, respectively. Since the exact evaluation protocol has some differences for each data set, we use the Pascal VOC 2007~\cite{pascal-voc-2007} metric for the detection experiments and the Cityscapes~\cite{Cordts2016Cityscapes} metric for the instance segmentation results. We use the evaluation code available at https://github.com/facebookresearch/maskrcnn-benchmark.  More details about the evaluation metrics can be found in the original papers~\cite{Cordts2016Cityscapes,pascal-voc-2007}.

\begin{figure*}[ht]
\vskip 0.2in
\begin{center}
\centerline{\includegraphics[width=\linewidth]{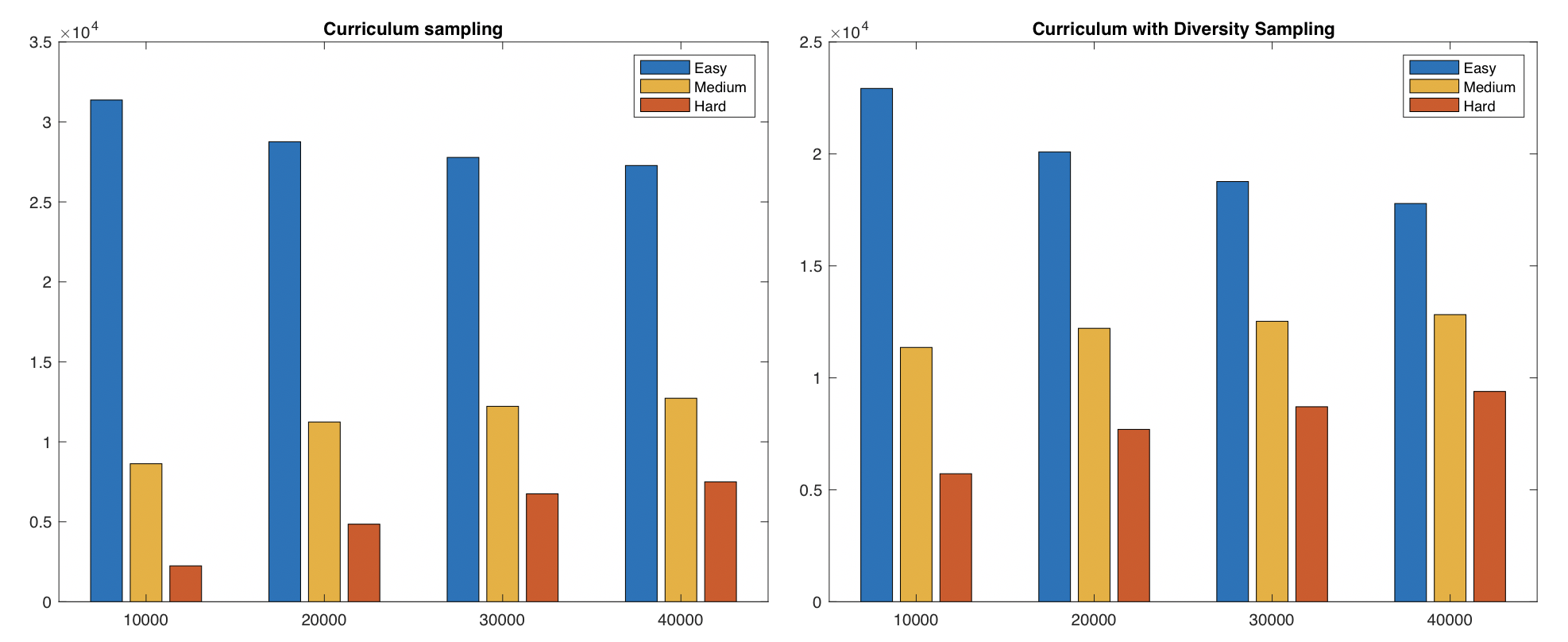}}
\caption{Difficulty of the images samples during our training on Pascal VOC 2007. On the left it is presented the curriculum sampling method and on the right the curriculum with diversity approach. We present the first 40000 iterations for each case, with histograms generated from 10k to 10k steps. Best viewed in color.}
\label{fig2}
\end{center}
\vskip 0.1cm
\end{figure*}

\subsection{Results and discussion}
The class distribution of the objects in Pascal VOC 2007 clearly favors class \textit{person}, with 4690 instances, while classes \textit{dinningtable} and \textit{bus} only contain 215 and 229 instances, respectively. This would not be a problem if the difficulty of the classes was similar, because we can assume the test data set has a matching distribution, but this is not the case, as it is shown in Figure~\ref{diff2}.

Figure~\ref{fig1} presents how the two sampling methods behave during training on the Pascal VOC 2007 data set. In the first 10k iterations, curriculum sampling selects images with almost 20k objects from class \textit{person} and only 283 instances from class \textit{diningtable}. By adding diversity, we lower the gap between classes, reaching 10k objects of persons and 1000 instances of tables. This behaviour continues as the training progresses, with the differences between classes being smaller when adding diversity. It is important to note that we do not want to sample the exact number of objects from each class, but to keep the class distribution of the actual data set, while feeding the model with enough details about every class. Figure~\ref{fig2} shows the difficulty of the examples sampled according to our strategies. We observe that by adding diversity we do not break our curriculum learning schedule, the examples still being selected from easy to hard.

To further prove the efficiency of our method, we compute the AP on both object detection and instance segmentation tasks. The results are presented in Tables~\ref{tab1} and~\ref{tab2}.

We repeat our object detection experiments five times and average the results, in order to ensure their relevance. The sampling with diversity approach provides an improvement of $0.69\%$ over the standard curriculum method, and of $0.79\%$ over the randomly-trained baseline. Although the improvement is not large, we can observe that by adding diversity we boost the accuracy where the standard method would fail, without much effort. Our experiments, with an inverse curriculum approach, from hard to easy, lead to the worst results, showing the utility of presenting the training samples in a meaningful order, similar to the way people learn. 

Moreover, Figure~\ref{fig3} illustrates the evolution of the AP during training. The curriculum with diversity approach has superior results over the baseline from the beginning to the end of the training. As the figure shows, the difference between the two methods increases in the later stages of the training. A simple reason for this behaviour is the fact that the curriculum strategy is fed with new, more difficult, examples as the training progresses, continuously improving the accuracy of the model. On the other hand, the standard random procedure receives all information from the beginning, reaching a plateau early during training. The standard CL method starts from lower scores, exactly because it does not visit enough samples from more difficult classes in the early stages of the training. For instance, after 5000 iterations, the AP of the standard CL approach on class \textit{dinningtable} was $0$. Thus, by adding diversity, our model converges faster than the traditional methods.
\begin{table}
\caption{Average Precision scores for object detection on Pascal VOC 2007 data set.}\label{tab1}
\centering
\begin{tabular}{|cc|}
\hline
Model & mAP (in \%)\\
\hline
Faster R-CNN (Baseline) & $72.28 \pm{0.34}$\\
Faster R-CNN with curriculum sampling & $72.38 \pm{0.32}$\\
Faster R-CNN with inverse curriculum sampling & $70.89 \pm{0.53}$\\
{\bf Faster R-CNN with diverse curriculum sampling} & $\mathbf{ 73.07 \pm{0.28}}$\\
\hline
\end{tabular}
\end{table}


\begin{table}
\caption{Average Precision scores for instance segmentation on Cityscapes data set.}\label{tab2}
\centering
\begin{tabular}{|cccc|}
\hline
Model & AP & AP50\% & AP75\%\\
\hline
Faster R-CNN (baseline) & $38.72$ & $69.15$ & $34.95$\\
Curriculum sampling & $38.47$ & {\bf 69.88} & $35.01$\\
Inverse curriculum sampling & $37.40$ & $68.17$ & $34.22$\\
Diverse curriculum sampling & {\bf 39.12} & $69.86$ & {\bf 35.4}\\
\hline
\end{tabular}
\end{table}

The instance segmentation results on the Cityscapes data set confirm the conclusion from our previous experiments. As Table~\ref{tab2} shows, the curriculum with diversity is again the optimal method, surpassing the baseline with $0.4\%$ using AP, $0.71\%$ using AP50\%, and $0.45\%$ using AP75\%. It is interesting to point out that, although the diverse curriculum approach has a better AP and AP75\% than the standard CL method, the former technique surpasses our method with $0.02\%$ when evaluated using AP50\%. The inverse curriculum approach has the worst scores again, strengthening our statements on the utility of curriculum learning and the importance of providing training examples in a meaningful order.

\section{Conclusion and future work}\label{sec_Conclusion}
In this paper, we presented a simple method of optimizing the curriculum learning approaches on unbalanced data sets. We consider that the diversity of the selected examples is just as important as their difficulty, and neglecting this fact may slow down training for more difficult classes. We introduced a novel sampling function, which uses the classes of the visited examples together with a difficulty score to ensure the curriculum schedule and the diversity of the selection. Our object detection and instance segmentation experiments conducted on two data sets of high interest prove the superiority of our method over the randomly-trained baseline and over the standard CL approach. A benefit of our methodology is that it can be used on top of any deep learning model, for any supervised task. Diversity can be a key element for overcoming one of the shortcomings of CL which can lead to the replacement of the traditional random training and a larger adoption of meaningful sample selection. For the future work, we plan on studying more difficulty measures to build an extensive view on how the chosen metric affects the performance of our system. Furthermore, we aim to create an ablation study on the parameter choice and find better ways to detect the right parameter values. Another important aspect we are considering is extending the framework to unsupervised tasks, by introducing a novel method of computing the diversity of the examples.

\bibliography{ecai}
\end{document}